\documentclass[letterpaper]{article} % DO NOT CHANGE THIS
\usepackage{aaai23}  % DO NOT CHANGE THIS
\usepackage{times}  % DO NOT CHANGE THIS
\usepackage{helvet}  % DO NOT CHANGE THIS
\usepackage{courier}  % DO NOT CHANGE THIS
\usepackage[hyphens]{url}  % DO NOT CHANGE THIS
\usepackage{graphicx} % DO NOT CHANGE THIS
\usepackage{natbib}  % DO NOT CHANGE THIS AND DO NOT ADD ANY OPTIONS TO IT
\usepackage{caption} % DO NOT CHANGE THIS AND DO NOT ADD ANY OPTIONS TO IT
\usepackage{algorithm}
\usepackage{algorithmic}
\usepackage{multirow}

\usepackage{newfloat}
\usepackage{listings}
\DeclareCaptionStyle{ruled}{labelfont=normalfont,labelsep=colon,strut=off} % DO NOT CHANGE THIS
\floatstyle{ruled}
\newfloat{listing}{tb}{lst}{}
\floatname{listing}{Listing}
\title{\#maskUp: Selective Attribute Encryption for Sensitive Vocalization for English language on Social Media Platforms}
\author{
    Supriti Vijay\textsuperscript{\rm 1}\equalcontrib,
    Aman Priyanshu\textsuperscript{\rm 2}\equalcontrib
}
\affiliations{
    \textsuperscript{\rm 1}Department of Computer Science\\
    \textsuperscript{\rm 2}Department of Information \& Communication Technology\\
    Manipal Institute of Technology,\\
    Manipal Academy of Higher Education,\\
    Karnataka, India.\\
    supriti.vijay@gmail.com
}

\usepackage{bibentry}
\begin{document}

\maketitle

\begin{abstract}
Social media has become a platform for people to stand up and raise their voices against social and criminal acts. Vocalization of such information has allowed the investigation and identification of criminals. However, revealing such sensitive information may jeopardize the victim's safety. We propose \#maskUp, a safe method for information communication in a secure fashion to the relevant authorities, discouraging potential bullying of the victim. This would ensure security by conserving their privacy through natural language processing supplemented with selective encryption for sensitive attribute masking. To our knowledge, this is the first work that aims to protect the privacy of the victims by masking their private details as well as emboldening them to come forward to report crimes. The use of masking technology allows only binding authorities to view/un-mask this data. We construct and evaluate the proposed methodology on continual learning tasks, allowing practical implementation of the same in a real-world scenario. \#maskUp successfully demonstrates this integration on sample datasets validating the presented objective.
\end{abstract}

\section{Introduction}

The rise in gender-based crimes has been alarmingly high over the past few years. Reports show that globally, 1 in 3 women experience physical and/or sexual violence in their lifetime \cite{WHO_2021}. However, the willingness of survivors to report such heinous crimes is very low \cite{article}. Less than 10 per cent of women who experience violence seek help from the concerned authorities. This lack of vocalization not only encourages criminals to harm again without fear but also allows such occurrences to continue and be prolonged. 

Societal and structural barriers like societal stigma and shame, distrust of institutions, fear of retaliation by the perpetrator, misuse of power by concerned authorities and prolonged trials, prevent women from coming forward and reporting crimes \cite{article}. Therefore, a platform to voice opinions without fear of societal judgment, devoid of misuse of power by institutions, is required to encourage women to speak up. Social media giants like Twitter, Facebook, Instagram, and Reddit have been very instrumental in being such a platform.

The \#MeToo movement marked a landmark year for the conversation about sexual abuse and violence on such platforms. It posed a method to voice one's opinions against societal suppression and garner support from those suffering under similar circumstances. Public activism emboldened victims of sexual abuse to come forward and speak out. Social media has become one of the strongest pillars for raising awareness, and fighting said crimes, allowing individuals to relay their support to those in delicate positions and seek the same. 

However, one must understand that information relayed on topics as volatile as Sexual Harassment and Crimes \cite{Koss1993RapeSI} can leak sensitive information and cause disastrous outcomes. Victims of sexual assault are often held culpable for the assault and face tremendous backlash and personal attacks \cite{suvarna-bhalla-2020-notawhore}. With the rise of such crimes, it is essential to devise a computational framework that can identify and prevent the online victimization of sexual assault survivors who choose to report the crime. In our construction of the problem statement, we aim to estimate and accurately retrieve such information and provide security for the vocalization of crimes. While previous literature and implementation focus on the identification of victim-blaming language and overall data encryption, they do not account for the impact these messages may have and the inherent fear amongst victims to come forward. It also does not account for how essential it is to communicate said information. 

Therefore, a methodology that only encrypts vital information that may be limited to characteristic names, locations, or other sensitive aspects of the texts is proposed. For this, we offer a streamlined pipeline that augments Named Entity Recognition with Selective Encryption to formalize Selective Attribute Encryption for Sensitive Vocalization on Social Media Platforms. To our knowledge, this is the first work in the field of computational social science that aims to protect the privacy of the victims by masking their confidential details.

\section{Related Work}

\subsection{Selective Encryption}

Selective encryption, a recently popularized field of Cryptography, proposes a trade-off between security and computational complexity. It is based on the constitution that encrypting only the sensitive aspects of the complete data gives enough encryption to conserve data privacy. Previous literature has proposed a multitude of methodologies that encrypt and secure texts, such as \cite{KUSHWAHA201616, Etaiwi2018}, which uses a symmetric-key-based encryption algorithm for selective encryption of text over a mobile ad hoc network. On the other hand, \cite{Kushwaha2018} uses natural language processing to optimize selective encryption for sensitive aspects of text sent over the same medium. However, these encryption systems are still computationally expensive, and although they retain their selective nature, the content may still be perceptible in some instances. Therefore, it becomes imperative that an updateable retrieval function be applied for sensitive attribute extraction.

\begin{figure*}[h!]
\begin{center}

  \includegraphics[width=0.8\textwidth,height=8cm]{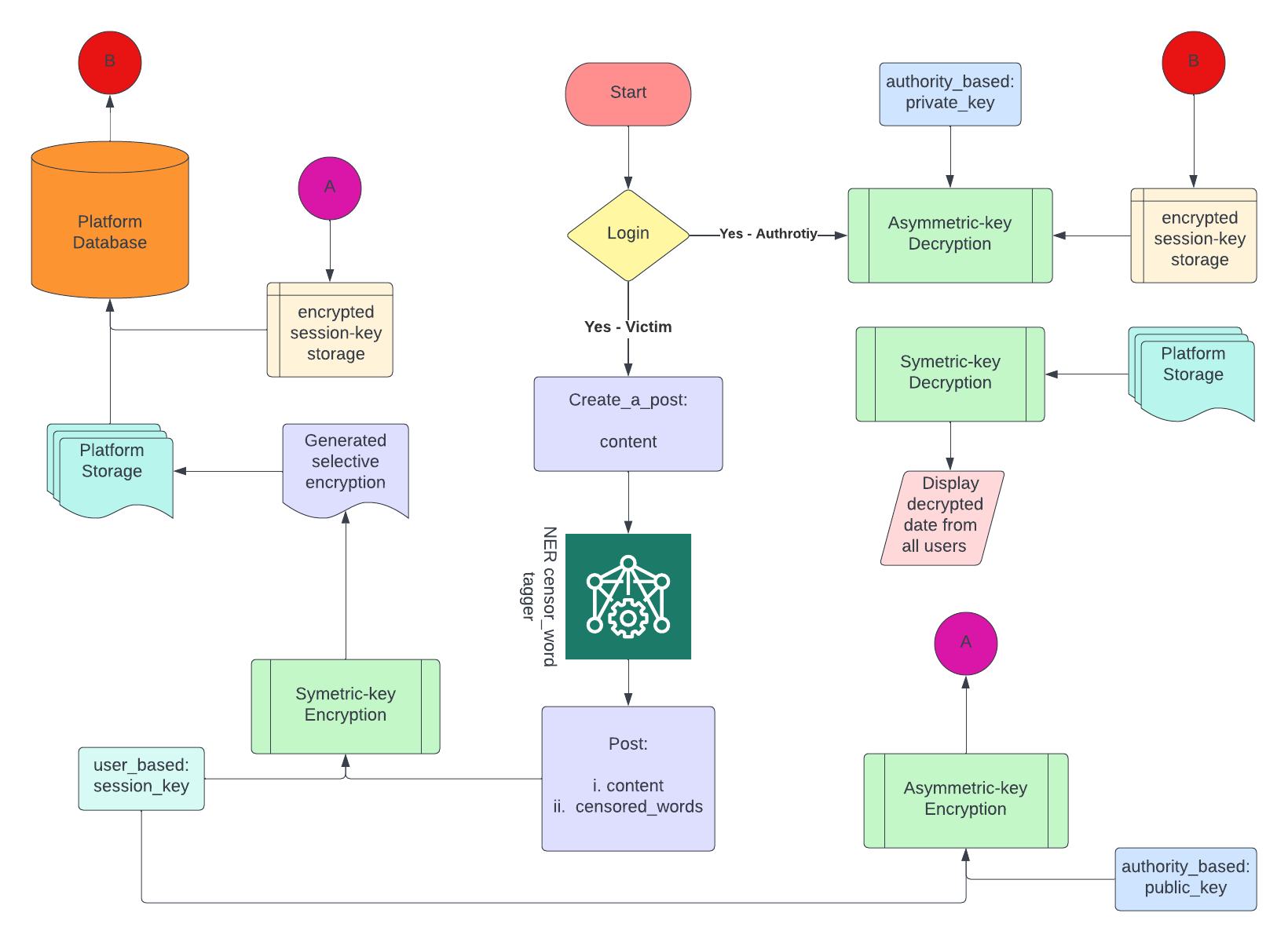}
  \caption{Algorithmic flow chart of \#maskUp detailing flow of data and encryption keys for given objectives. Included are the (1) Sensitive-Entity-Retrieval System (2) User Based Symmetric-Key Encryption Mechanism (3) Master-Key Module. Here, \(A\) and \(B\) simply act as connectors co-joining Asymmetric-key Encryption with encrypted session-key storage and Platform Database with encrypted session-key storage.}
  \label{fig:1}
  \end{center}
\end{figure*}

\subsection{Named Entity Recognition}
Named entity recognition has been an effective way of identifying and classifying names of person (PER), location (LOC), organization (ORG). Research on NER started with the use of handcraft features \cite{zhou-su-2002-named,chieu-ng-2002-named, 10.3115/1119176.1119196, 10.5555/1567594.1567618}, supervised learning techniques \cite{https://doi.org/10.48550/arxiv.2101.11420}  joint structured CRF models \cite{durrett-klein-2014-joint} and transitioned to semi-supervised learning methods \cite{10.1007/11766247_23, 10.3115/1072228.1072382, 10.5555/315149.315364,cucchiarelli-velardi-2001-unsupervised,10.5555/1597348.1597411} due to limited structured data.

Recently, deep learning methods came to light since they significantly showed progress by automatically extracting high-level features and performing sequence tagging with neural networks \cite{https://doi.org/10.48550/arxiv.1505.05008, https://doi.org/10.48550/arxiv.1511.08308, https://doi.org/10.48550/arxiv.1603.01360, https://doi.org/10.48550/arxiv.1910.11470}. The rise of transformers brought promising results for, \cite{https://doi.org/10.48550/arxiv.1911.04474} proposed TENER, a NER architecture adopting an adapted Transformer Encoder to model the character-level features and word-level features. Thus, utilizing the best technique for NER, in our proposed algorithm, we use the NERDA Framework, an open-sourced tool that fine-tunes transformers for NER tasks for any arbitrary language to identify sensitive information from text.

\subsection{Continual Learning}

Continual learning refers to the ability to continually learn over time by accommodating new knowledge while retaining previously learned experiences \cite{PARISI201954}. Research in this field has found significant development by utilising the concept of regularisation. Joint learning, an implementation accommodating the procedures of continual learning, requires interleaving samples from each task \cite{Caruana1997}. However, this methodology becomes increasingly cumbersome as the number of tasks increases making it resource-constraining.

This led to the development of Learning without Forgetting ($LwF$) \cite{li2017learning}, which only required samples form the task-at-hand for learning. Although similar to joint training, the method does not require old data or reference points; instead, it uses regularisation techniques such as, Elastic Weight Consolidation, to compensate for the network forgetting an entire sequence of old data.

By using the $LwF$ algorithm, the model is able to retain the previous performance learnt on the old tasks as well as gradually learn and update for newer tasks. However, this methodology still does not account for gradient flow and may restrain the training over the current task. Therefore, we restrict the training only to those neurons which have shown value in previous tasks. This idea was proposed in the "Overcoming catastrophic forgetting in neural networks" paper which discusses the integration of the Fisher information matrix as a regularizer for the loss function \cite{ewc_paper}. We provide said equation \ref{equ:1} and integrate it into the training of our proposed \#maskUp algorithm.

\begin{equation}
\label{equ:1}
  {\cal L}(\theta) = {\cal L}_B(\theta) + \sum_i \frac{\lambda}{2} F_i (\theta_i - \theta^*_{A,i})^2
\end{equation}

where $ {\cal L_B}(\theta) $ is the loss for task B only, $\lambda$ defines how important the old task is compared to the new one and $i$ labels each parameter.

With the introduction of continual learning, we add a dimension of adaptability and robustness against domain shifts for our algorithm.

\section{Proposed Methodology}

We consider our primary problem statement to be the construction of an NER based sensitive information encryption system. The retrieval system must incorporate privacy leaking terms such as names of individuals involved in the incident, as well as, any locations/organizations that may be used to retrace or reconstruct user identity. We present our algorithmic flowchart in Fig~\ref{fig:1}. Through \#maskUp, we aim to achieve the following objectives:

\begin{enumerate}
    \item Extract sensitive phrases/words from given paragraph/post/tweet (further referred to as a document).
    \item Enable Continual Learning using $EWC$ paradigm for targeted neuron training.
    \item Selectively encrypt only those aspects of the document which may be sensitive to the users (such as names, locations, organizations, among others.).
    \item Provide a mechanism for criminal authorities to decrypt all user data (Master-Key Module).
\end{enumerate}

We demonstrate a flow of data and encryption keys through our proposed methodology.

\subsection{Continual NER}

For the active and continual deployment of \#maskUp, we recognized the need for integration with online learning paradigms. As language and grammatical structures tend to shift, it becomes imperative for said model to incorporate and augment its parameters with this new data.

Named Entity Recognition becomes an important aspect of our proposed methodology as it seeks to retrieve those entities which may leak victim privacy. Our objective aims to fine-tune an NER for the task of sensitive feature retrieval. We train the NER on the CoNLL-2003 Dataset, a benchmark in NER tasks. The CoNLL-2003 is a named entity recognition dataset released as a part of CoNLL-2003 shared task: language-independent named entity recognition. Following are the entities the model learns to distinguish with respect to said dataset,
\begin{table}[]
\begin{center}
\begin{tabular}{llll}
• B-PER & • I-PER & • B-ORG & • I-ORG \\ \\
• B-LOC & • I-LOC & • B-MISC & • I-MISC
\end{tabular}
\end{center}
\end{table}
The dataset follows a prescribed template where the words tagged with $O$ are outside of the named entities while the $I-XXX$ tag is used for words inside a named entity of type $XXX$. Whenever two entities of type $XXX$ are immediately next to each other, the first word of the second entity will be tagged $B-XXX$ in order to show that it starts another entity. The data includes entities of four types: persons ($PER$), organizations ($ORG$), locations ($LOC$) and miscellaneous names ($MISC$) as mentioned above. Each of which are relevant to our extraction of sensitive attributes.

This enables the model to retrieve only those aspects of the dataset which may include personal details, such as names, associated organizations,locations, among other sensitive attributes. Utilizing the depth and variety of CoNLL-2003 dataset, allows us ensurance over our collection.

We aim to integrate a rectification feature, whereby users may choose to use/not use certain words selected by them, allowing our models to utilize the concept of $EWC$ for progressive learning overtime \cite{ewc_paper}. This corrective feature would further allow users to include their own words, making the model synchronous with the users' linguistics.

\subsection{Selective Attribute Encryption}

\begin{table*}[]
\begin{center}
\begin{tabular}{|l|l|l|l|}
\hline
\textbf{Algorithm Name} & \textbf{Time Taken - Encryption} & \textbf{Time Taken - Decryption} & \textbf{Memory Utilized} \\ \hline
Full AES Encryption & 2038.5 & 1970.4 & 1.406 \\ \hline
\#maskUp & 211.6 & 207.5 & 0.553 \\ \hline
\end{tabular}
\caption{\label{table:2} Performance of Complete AES Encryption against Selective AES Encryption. Relayed Time Taken in ms and Memory Utilization (kB). Performance generalized over 30 articles in each instance of comparison.}
\end{center}
\end{table*}

Selective attribute encryption is enabled as the base mechanism for privacy-preservation within \#maskUp. We employ AES---Symmetric Key Encryption system for securing said data. The symmetric-key is generated for every user independent of others, it utilizes their given $PLATFORM\_PASSWORD$ and employs a $NOISE\_TRANSFORMATION$ coupled with $DOUBLE\_ENCRYPTION$ of the output, generating a key of $KEY\_SIZE=128 bits$. This key is then utilized for selective encryption of target entities within the user text. The rest of the document is returned as user-provided and no changes are made to it. We choose to employ AES for the encryption mechanism due to its wide usage, as well as its ingrained security.

\begin{table}[!ht]
    \centering
    \resizebox{\columnwidth}{!}{\begin{tabular}{|l|l|l|l|}
    \hline
        \textbf{Level} & \textbf{F1-Score} & \textbf{Precision} & \textbf{Recall} \\ \hline
        \textbf{B-PER} & 0.949689 & 0.953242 & 0.946163 \\ \hline
        \textbf{I-PER} & 0.984483 & 0.980258 & 0.988745 \\ \hline
        \textbf{B-ORG} & 0.868681 & 0.890082 & 0.848284 \\ \hline
        \textbf{I-ORG} & 0.839024 & 0.854658 & 0.823952 \\ \hline
        \textbf{B-LOC} & 0.914252 & 0.901458 & 0.927415 \\ \hline
        \textbf{I-LOC} & 0.811808 & 0.769231 & 0.859375 \\ \hline
        \textbf{B-MISC} & 0.807584 & 0.796399 & 0.819088 \\ \hline
        \textbf{I-MISC} & 0.64488 & 0.609053 & 0.685185 \\ \hline
        \textbf{AVG\_MICRO} & 0.894215 & --- & --- \\ \hline
        \textbf{AVG\_MICRO} & 0.85255 & --- & --- \\ \hline
    \end{tabular}}
    \caption{\label{table:1} Performance of 5-Fold Cross Validation of Google's electra-small-discriminator model for NER on the CoNLL-2003 dataset.}
\end{table}

\subsection{Master-Key Module}

While the utilization of symmetric key encryption provides validity and safety in securing user data, it does not account for visibility to civic authorities. It becomes important that legally binding authorities understand and evaluate concerns regarding said posts, and therefore, a Master-Key module is created which makes use of asymmetric-key encryption system. We specifically employ RSA-encryption for encrypting the symmetric keys of every user. The platform utilizes $PUBLIC_KEY$ provided by the legally binding authority of that country/state. A $PRIVATE_KEY$ is used by the authorities to decrypt the said symmetric key. Since the authorities aren't aware of the $NOISE\_TRANSFORMATION$ coupled with $DOUBLE\_ENCRYPTION$ mechanism, we ensure the privacy of User passwords. This ability acts as a master key, allowing users to easily navigate their own accounts independently of other users. However, authorities can overlook and decrypt sensitive attributes from whichever account they wish to pull.

\section{Result Analysis}

For our implementation, we incorporate the $NERDA$ library for NER-model training, we specifically utilize the "google/electra-small-discriminator" model for its state-of-the-art electra implementation as well its comparatively smaller base. We provide in Table~\ref{table:1} its training over the CoNLL-2003 corpora.

We provide a comparative analysis in Table~\ref{table:2} for complete encryption and selective encryption of data by \#maskUp. The execution time was averaged over 30 distinct articles giving us a generalized overview of time-memory utilization.

The experiments clearly distinguish performance of selective encryption from complete encryption. With an optimization of upto 89.1\% for time-taken during encryption and 60.7\% for memory consumed, \#maskUp considerably reduces deployment costs on edge-devices. The further integration of continual learning also supplements the privacy-preserving features of this proposal. Our utilization of $EWC$ is inspired by the $NERDA-Con$ python library \cite{https://doi.org/10.48550/arxiv.2206.14607}, which is a pipeline for training NERs with LLM bases by incorporating the concept of Elastic Weight Consolidation ($EWC$) into the NER fine-tuning NERDA pipeline.

\section{Ethical Considerations}

We work with the aim to help vocalize victims of gender-based crimes on social media to expedite the process of seeking assistance from the concerned authorities. We acknowledge that sensitive information may be subjective and open for interpretation. We take further precautions to ensure data doesn't get leaked and is directly accessible to only authorities who possess the master key. However, we recognize that it is almost impossible to prevent abuse of released technology even when developed with good intentions \cite{hovy-spruit-2016-social}. All the examples shown in this paper have been taken from online reports, ensuring anonymization and paraphrasing for user privacy. We further acknowledge that our named entity recognition pipeline may be susceptible to allocation bias, regularity bias and bias against certain demographics\cite{ghaddar-etal-2021-context,https://doi.org/10.48550/arxiv.2008.03415}. However, the essence of our work is to create a safe space for abuse victims to come forward and voice their incidents directly to the authorities without having to fear the spread of information or societal stigma. Furthermore, it is essential that the authorities aren't overburdened by falsified complaints that hinder the road to justice. 

\section{Conclusion}
With a motivation to provide a safe space for victims of gender-based crimes and for authorities to investigate the identification of said criminals, we present \#maskUp, a method for information to be conveyed securely to the relevant authorities. \#maskUp utilizes masking technology to selectively encrypt sensitive information and continuously train on new data using continual learning. Sampled datasets are used to validate and ensure the working of the same. 
Future work can try to focus on the multilingual aspect of vocalization as well to include victims of all languages. We believe this method will help embolden victims to come forward and report incidents without any fear of societal judgement or misuse of power. 

\section*{Acknowledgements}

The authors of the paper are grateful to the reviewers for reviewing the manuscript and their valuable inputs are appreciated. We would also like to
thank the Research Society MIT for supporting the
project.

\bibliography{aaai23}

\appendix

\begin{figure*}[h!]
\begin{center}
  \includegraphics[width=0.9\textwidth,height=11.5cm]{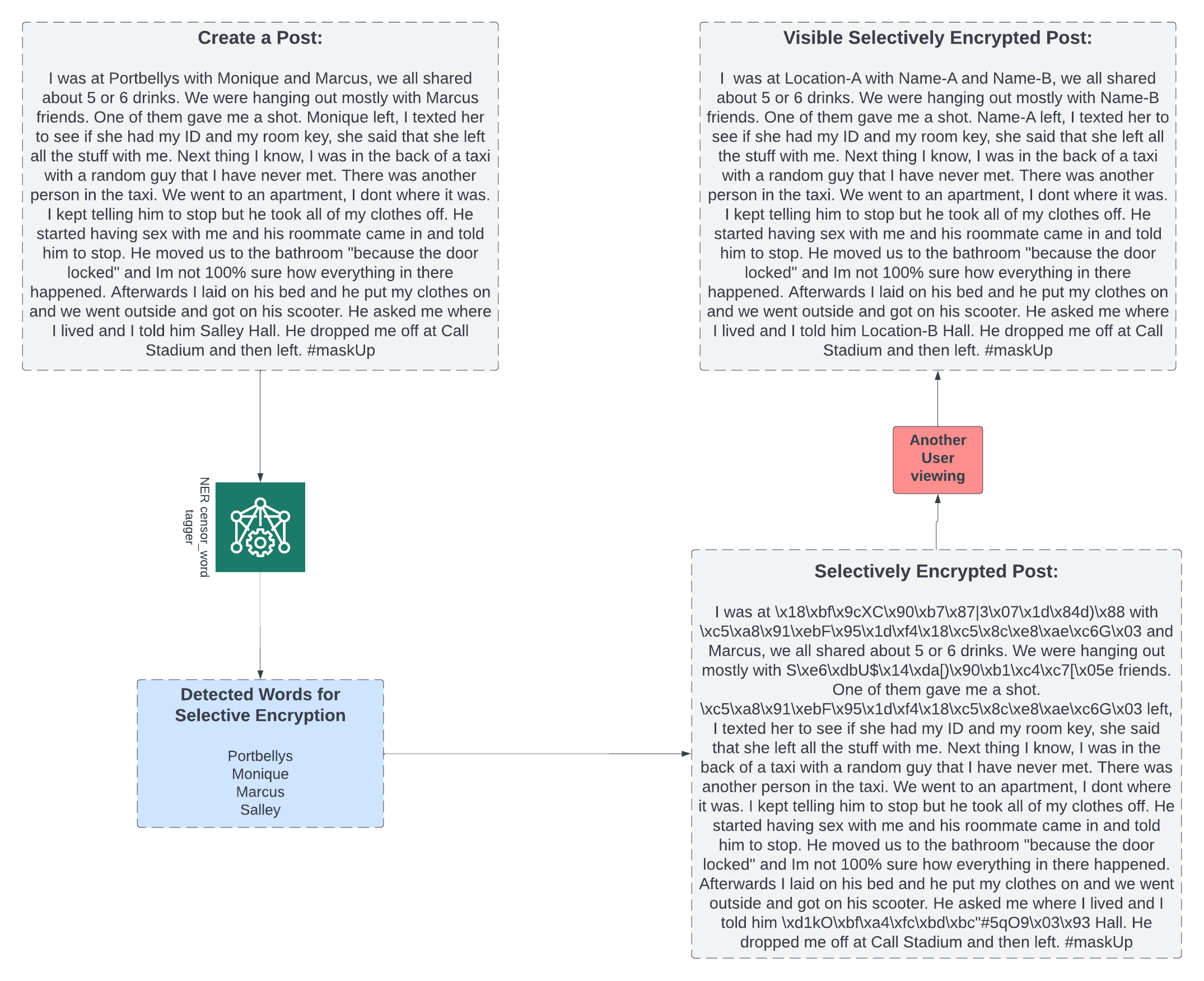}
  \caption{Algorithmic flow of a User's post through \#maskUp.}
  \label{fig:2}
  \end{center}
\end{figure*}

\end{document}